\documentclass[11pt,a4paper]{article}
\usepackage[T1]{fontenc}
\usepackage[utf8]{inputenc}
\usepackage{lmodern}
\usepackage[a4paper,margin=23mm,headheight=14pt]{geometry}
\usepackage{microtype}
\usepackage{amsmath,amssymb}
\usepackage{booktabs,tabularx,array}
\usepackage{enumitem}
\usepackage{graphicx}
\usepackage{xcolor}
\usepackage{tikz}
\usetikzlibrary{arrows.meta,positioning,calc}
\usepackage[numbers,sort&compress]{natbib}
\usepackage{fancyhdr}
\usepackage{titlesec}
\usepackage{needspace}
\usepackage{hyperref}
\hypersetup{colorlinks=true,linkcolor=black,citecolor=black,urlcolor=blue!45!black,
  pdftitle={Intrinsic Robot Rewarding: Reusing VLA Representations for Autonomous Evaluation and Policy Improvement in Industrial Manipulation},pdfauthor={Tobias Schaffer},
  pdfsubject={A research position on autonomous robot evaluation and learning}}
\setlist{nosep,leftmargin=*}
\setlist[enumerate]{itemsep=4pt,topsep=5pt}
\titleformat{\section}{\large\bfseries}{\thesection}{0.7em}{}
\titleformat{\subsection}{\normalsize\bfseries}{\thesubsection}{0.7em}{}
\titlespacing*{\section}{0pt}{13pt}{6pt}
\titlespacing*{\subsection}{0pt}{9pt}{4pt}
\newcolumntype{Y}{>{\raggedright\arraybackslash}X}
\newcolumntype{L}[1]{>{\raggedright\arraybackslash}p{#1}}

\newcommand{\R}{\mathbb{R}}

\newcommand{\norm}[1]{\left\lVert#1\right\rVert}

\title{Intrinsic Robot Rewarding: Reusing VLA Representations for Autonomous Evaluation and Policy Improvement}

\author{
Tobias Schaffer\textsuperscript{1,*},
Mohab Elkhayat\textsuperscript{1},
Daniela Nicklas\textsuperscript{1},\\
Mustafa Almohamad\textsuperscript{1},
Elham Al-Fuqara\textsuperscript{1}
}

\date{}

\begin{document}
\maketitle

\begin{center}
\small
\textsuperscript{1}Technology Campus Cham - Intelligent Robotics,\\
Deggendorf Institute of Technology, Cham, Germany\\
\textsuperscript{*}Corresponding author: tobias.schaffer@th-deg.de
\end{center}

\vspace{0.5pt}

\begin{abstract}
Vision-language-action (VLA) systems already bring together two valuable resources for robot learning: rich visual representations and demonstrations of successful task execution. Intrinsic Robot Rewarding (IRR) proposes to use these resources for a second, complementary purpose: evaluating the robot's own outcomes and providing feedback for policy improvement. Successful demonstration endpoints define task-specific references, and the policy's frozen visual encoder provides the feature space in which new outcomes are assessed. The core reward mechanism adds a reference bank and a scoring operation to the existing pipeline, without requiring a separate learned evaluator or an additional perception backbone. Our position is that this reuse offers a promising route to lower integration effort, efficient reward computation, and reduced recurring human outcome scoring. Building on established research in visual rewards and learning from experience, IRR brings these ideas into the robot's existing perception and demonstration pipeline. An operational COMAU Racer 3 demonstrator is available at technology readiness level 4 (TRL 4). This laboratory foundation supports the next research step: connecting internal outcome evaluation to physical policy improvement. We present the reward formulation, central research questions, and an evaluation methodology linking reward reliability to task success and supervision effort. The intended contribution is a reusable approach to learn and improve from the data and experience already available in industrial robot systems.
\end{abstract}

\section{Position and motivation}
\label{sec:position}

Vision--language--action models connect visual observations and language instructions to robot actions. RT-2 demonstrated the transfer of web-derived semantic knowledge into robotic control, while OpenVLA made a broadly pretrained VLA available for downstream adaptation \citep{rt2,openvla}. These developments provide an increasingly capable foundation for manipulation. Adapting such a policy to an industrial task also produces a valuable local resource with demonstrations that show both, how the task is performed and what successful execution looks like.

We take the position that \emph{the representations and successful examples already available in a robot's learning pipeline are a valuable resource for evaluating its own performance}. A demonstration contains more than action supervision. Its endpoint provides evidence of the intended outcome, and the policy's visual encoder already offers a representation in which that outcome can be described. Intrinsic Robot Rewarding puts these resources to work together: successful endpoints form a task-specific reference bank, and new attempts are scored by their relationship to those references.

The central benefit of this design is reuse. In its core form, IRR adds reference management and similarity scoring to the existing perception pipeline. It requires no separate reward foundation model, no additional perception backbone, and no new reward-model training corpus for constructing the reference bank. This offers a practical opportunity to reduce integration and model-maintenance effort while making fuller use of demonstrations that have already been collected. Where reward and policy inputs coincide, visual features can also be shared at inference time. The magnitude of these benefits will be measured through computation, engineering effort, and human supervision.

An intuitive analogy is learning to stack or balance objects. A learner may recognize that a stack has collapsed, or that an object has lost its balance, before being able to perform the task reliably. Recognizing the outcome and producing it consistently are related but distinct capabilities. Perceiving the consequences of an attempt provides feedback for adjusting the next one. IRR adopts this distinction at the system level: a robot's existing visual representation may already contain information useful for recognizing successful outcomes, even while its action policy is still learning to achieve them consistently. This motivates the use of shared perceptual resources for both acting and evaluating.

Reinforcement learning (RL) provides the mechanism for turning such outcome feedback into improved behavior. Geometric rewards, human outcome judgments, and dedicated vision--language model (VLM) evaluators are established ways to support this learning. IRR contributes an additional design option: obtaining the reward from resources already integrated into the task policy. This is particularly relevant when industrial systems must be adapted repeatedly to new parts, destinations, fixtures, or workspace conditions.

Here, \emph{intrinsic} refers to reward computation within the robot's existing perception and learning pipeline. The task remains specified by people through instructions and successful demonstrations. IRR is therefore a task-conditioned, demonstration-derived outcome reward while curiosity approaches instead reward novelty or prediction error \citep{pathak2017}. The intended increase in autonomy concerns repeated outcome evaluation and feedback during learning. Commissioning, independent validation, and recovery remain visible parts of the overall effort.

Our available VLA demonstrator and its documented laboratory results provide a concrete foundation for this direction \citep{pipeline}. This paper develops the scientific position, the reward formulation, and the evidence needed to assess representation quality, physical learning, efficiency, and transfer. The reported measurements describe the existing imitation-trained system while improvements obtained through IRR are the subject of the proposed research. Together, the operational foundation and the principle of reuse establish a practical path toward self-evaluating and self-improving industrial robots.

\section{Related research and proposed contribution}
\label{sec:related}

\subsection{Policy adaptation and learning from experience}

OpenVLA combines DINOv2 and SigLIP visual features with a language backbone and action prediction~\citep{openvla,dinov2,siglip}. Its optimized fine-tuning recipe, OpenVLA-OFT, uses parallel decoding, action chunking, continuous actions, and an L1 regression objective~\citep{oft}. This provides a practical base policy for studying internal outcome evaluation. A residual RL controller can build on its regression-based action interface while supplying the stochastic learning mechanism for subsequent improvement.

Progress in real robot learning gives further support to this direction. HIL-SERL integrates demonstrations, human corrections, and efficient RL for dexterous manipulation~\citep{hilserl}. The $\pi_{0.6}^{*}$ system and RECAP combine demonstrations, experience, and corrections in VLA learning~\citep{recap}. Together, these approaches motivate the transition from imitation to learning through task experience. IRR focuses on making the associated reward provision easier to integrate by drawing on the policy's existing perceptual representation.

\subsection{Visual rewards and goal representations}

Value-implicit pre-training (VIP) learns a visual representation in which distance to a goal image can define rewards for downstream robot tasks~\citep{vip}. LIV jointly learns language--image representations and rewards from video with text annotations~\citep{liv}. RoboCLIP uses a pretrained video--language model to produce reward from a video or text demonstration~\citep{roboclip}. Baumli et al. derive rewards for visual language goals from pretrained VLMs~\citep{vlmrewards}. The listed research provides a substantial foundation for our IRR proposition that learned visual representations can support useful task feedback.

IRR develops this around the resources of an existing VLA deployment. Its contribution centers on the joint use of the policy encoder and the demonstrations already collected for imitation learning, with task-specific reference construction and evaluation on industrial hardware. The relevant scientific question is how effectively this reuse supports reward generation and policy improvement, and what integration and supervision benefits it will provide.

RL-VLM-F obtains preference feedback from a VLM and learns a reward function~\citep{rlvlmf}. GoalLadder incrementally discovers and ranks goal states using VLM comparisons, then trains an agent to approach a selected goal in a learned embedding space~\citep{goalladder}. These methods demonstrate complementary ways to organize goal information and feedback. IRR investigates a directly available source of such information, the successful endpoints from the task's existing demonstrations, represented by the policy's own visual encoder.

\subsection{Self-referential rewards and recent reward models}

SRPO is closely related: it uses successful trajectories from the current rollout batch as references and latent representations from V-JEPA~2 to provide progress rewards for unsuccessful trajectories~\citep{srpo,vjepa2}. Its formulation retains a sparse terminal success signal for identifying reference successes, and its reported physical experiments use an offline RL approach. IRR addresses a complementary question of how successful demonstration endpoints and the existing policy encoder can supply outcome feedback for repeated online physical learning. This connection places internal reference-based evaluation within an active and promising line of research.

Recent reward models also provide useful methodological foundations and comparators. RoboReward introduces a real robot reward dataset, benchmark, and VLM reward models, including negative and near-miss examples~\citep{roboreward}. Large Reward Models studies foundation VLMs adapted to generate process, completion, and temporal contrastive rewards for online refinement~\citep{largereward}. Their approaches inform the evaluation of outcome discrimination and learning utility. Comparisons with these models will establish where encoder reuse provides an attractive balance of task performance, computation, and supervision effort.

Table \ref{tab:positioning} summarizes the research position. The bibliography reflects selected primary literature and focuses on the methods most relevant to the proposed direction.

\begin{table}[t]
\caption{Relationship to closely related reward approaches and the contribution investigated through IRR.}
\label{tab:positioning}
\small
\begin{tabularx}{\linewidth}{@{}L{2.8cm}YY@{}}
\toprule
Approach & Reward source or reference & IRR research distinction \\
\midrule
VIP / LIV~\citep{vip,liv} & Representations trained to support visual or language goal rewards & Can the existing policy representation suffice without reward-oriented pretraining? \\
RoboCLIP~\citep{roboclip} & Video/text reference and pretrained video--language encoder & Can the same imitation data and policy encoder serve both action and outcome evaluation? \\
RL-VLM-F / GoalLadder~\citep{rlvlmf,goalladder} & VLM feedback for reward learning or goal ranking & Can task references reduce recurring external model feedback? \\
SRPO~\citep{srpo} & Successful rollouts, sparse success signal, and world-model trajectory features & Can an internal visual reference supply terminal reward for online physical learning? \\
RoboReward / Large Reward Models \citep{roboreward,largereward} & Separately trained VLM reward models & What reliability and cost trade-off does encoder reuse achieve? \\
\bottomrule
\end{tabularx}
\end{table}

\section{Available TRL 4 demonstrator and experimental foundation}
\label{sec:baseline}
\begin{figure}[htbp]
\centering
\includegraphics[width=\textwidth,height=5.5cm,keepaspectratio]{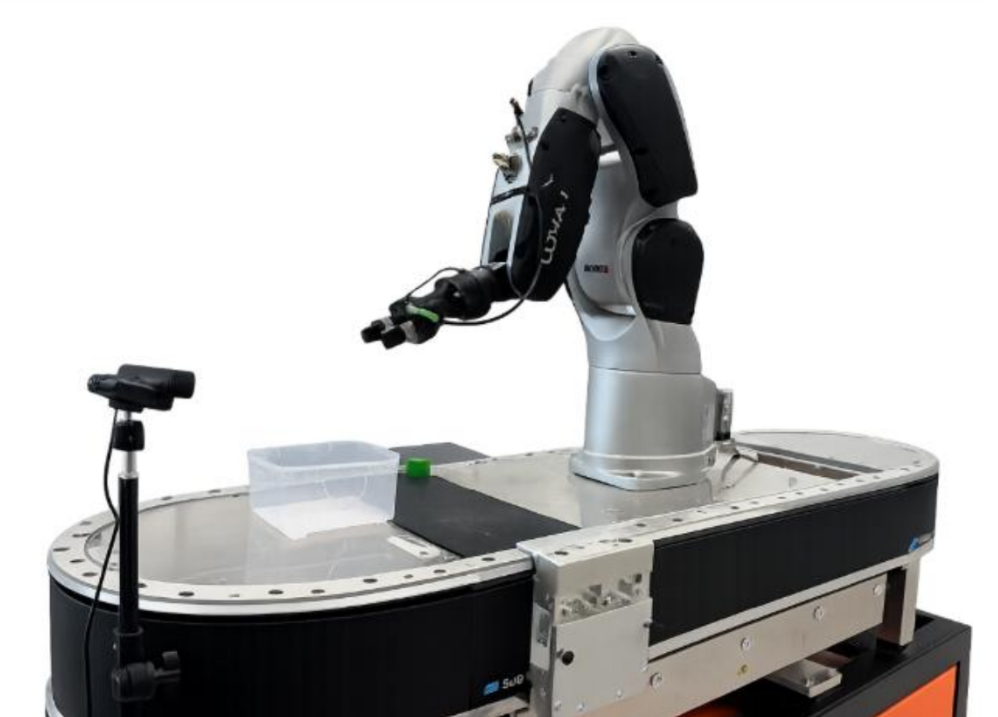}
\caption{\small Photograph of the available COMAU Racer 3 laboratory cell. The setup displays the integrated Robotiq Hand-E gripper, the primary camera viewpoint, the sample cube, and the designated target container.}
\label{fig:demonstrator}
\end{figure}

An operational VLA demonstrator is available at TRL~4, providing a laboratory-validated experimental foundation for IRR. The demonstrator comprises a COMAU Racer~3 six-axis industrial arm with a Robotiq Hand-E gripper, a third-person USB camera, VR teleoperation, and a ROS control stack~\citep{pipeline}. The baseline study uses 245 demonstrations of placing a small green cube into a container. Images, proprioceptive observations, and end-effector delta commands were recorded at a nominal 5\,Hz. A seven-component action comprises six Cartesian pose increments and a gripper command. OpenVLA-7B was adapted using LoRA and the OFT recipe, the reported experiment used five-action predictions and executed two actions before querying again. Figure~\ref{fig:demonstrator} shows a photograph of the setup.

The baseline experimental evaluation consists of 50 physical trials, each with a 60-second limit, where object and container placements are varied according to a fixed protocol. The aggregate results are shown in Table~\ref{tab:pilot}. The largest individual failure category was failure to center the end effector on the object, accounting for 8 of the 22 failed trials. This identifies a concrete opportunity for learning from outcome feedback, particularly through corrections to approach and placement behavior. Comparisons with observation and demonstration improvements will establish the specific contribution of reward-driven learning.

\begin{table}[ht]
\centering
\caption{Reported laboratory results of the available VLA demonstrator~\citep{pipeline}, providing the imitation-policy baseline for subsequent IRR research.}
\label{tab:pilot}
\small
\begin{tabularx}{0.95\linewidth}{@{}Yr@{}}
\toprule
Measure & Reported value \\
\midrule
Demonstration episodes & 245 \\
Successful evaluation trials & 28 / 50 (56\%) \\
Failures due to the time limit & 20 / 50 \\
Failures involving singularity / emergency stop & 2 / 50 \\
Completion time among successful trials & $39.16\pm9.50$\,s (mean $\pm$ SD) \\
Approximate 95\% Wilson interval for 28 / 50\textsuperscript{*} & 42.3 \ldots 68.8\% \\
\bottomrule
\end{tabularx}
\vspace{3pt}
\begin{minipage}{0.95\linewidth}\footnotesize
\textsuperscript{*}Calculated here from the aggregate count under a binomial model. 
\end{minipage}
\end{table}

The TRL~4 demonstrator integrates the robot, teleoperation, demonstration collection, VLA adaptation, and closed-loop execution in a working laboratory system. Its 56\% task success rate provides both an established baseline and a measurable opportunity for further learning. The proposed research can therefore build directly on available hardware, data, and software. Matched comparisons will use a newly measured baseline with the policy checkpoint, observations, control settings, and trial conditions held fixed.

\section{Proposed IRR formulation}
\label{sec:method}
\subsection{Task-conditioned success references}
Let $g$ be a task with a written success specification, $o_t$ denote its camera observation, $p_t$ the robot's proprioceptive state, and $\pi_0$ the imitation-trained VLA policy. Let $\phi$ be a frozen checkpoint of its visual encoder and $P$ a documented feature extraction and pooling operation. For a single image, define
\begin{equation}
 z_t=\frac{P(\phi(o_t))}{\norm{P(\phi(o_t))}_2+\varepsilon}\in\R^d,
 \qquad \varepsilon>0.
 \label{eq:embedding}
\end{equation}
The encoder layer, image preprocessing, token pooling, and normalization are part of the method and must be fixed before evaluation. DINOv2 features, SigLIP features, and their fusion will be compared. A global pooled embedding is the simplest baseline.

Successful endpoints from the imitation-training demonstrations define
\begin{equation}
 \mathcal{Z}^{+}_{g}=\{z_{T_i}^{(i)}:\tau_i\in\mathcal{D}^{+}_{g,\mathrm{train}}\}.
 \label{eq:bank}
\end{equation}
Here $\tau_i$ is a demonstration trajectory and $T_i$ its endpoint. The task instruction also selects the appropriate bank (a visually valid endpoint for one task must not automatically reward another). A bank represents multiple valid outcomes, such as different acceptable object poses, rather than forcing them into one visual prototype. Separate episodes and collection sessions are reserved for calibration and final testing.

\subsection{Similarity as a candidate reward}
For $1\leq k\leq|\mathcal{Z}^{+}_{g}|$, let $\mathcal{N}_k(z;\mathcal{Z}^{+}_{g})$ be the $k$ nearest reference embeddings. A simple reference distance and score are
\begin{align}
 d_g(z)&=\frac{1}{k}\sum_{u\in\mathcal{N}_k(z;\mathcal{Z}^{+}_{g})}\norm{z-u}_2^2,
 \label{eq:distance}\\
 s_g(z)&=\exp[-d_g(z)/\tau_g],\qquad \tau_g>0.
 \label{eq:score}
\end{align}
In Equation~\eqref{eq:score}, $\tau_g$ is the positive, task-specific distance-scale (temperature) parameter. It controls how quickly the score decreases with distance from successful reference states, smaller $\tau_g$ makes the score more selective, whereas larger $\tau_g$ gives a more gradual decay. The parameter uses the same distance scale as $d_g$ and is distinct from the demonstration trajectory $\tau_i$ in Equation~\eqref{eq:bank}. Its value is selected from training or held-out calibration data, documented for each task, and fixed before learning and final evaluation.

The score lies in $(0,1]$ but is not a calibrated probability of success. The initial experiment will use a terminal reward, where \(T\) denotes the final time step of the attempt. To limit transient visual matches, an optional persistence score takes the minimum of the scores over a short, fixed observation window after the attempt:
\begin{equation}
 s^{\mathrm{pers}}_g=\min_{j=0,\ldots,m-1}s_g(z_{T-j}),\qquad
 r_T^{\mathrm{IRR}}=q_T s^{\mathrm{pers}}_g,\qquad r_t^{\mathrm{IRR}}=0\;(t<T).
 \label{eq:reward}
\end{equation}
The single-frame version has $m=1$. The validity indicator $q_T$ rejects missing or stale observations and protocol-invalid attempts, an independent safety abort also sets reward to zero and terminates the attempt. Multiple views and ordered terminal sequences will be investigated when a single view cannot distinguish success from a near miss.

\subsection{Calibration and reliable autonomous evaluation}

The core IRR reference model is constructed only from successful demonstrations already collected for imitation learning. For operational success detection, a task-specific threshold $\eta_g$ is applied to the persistence score:
\begin{equation}
\hat y =
\begin{cases}
1, & s^{\mathrm{pers}}_g \geq \eta_g \quad \text{(predicted success)},\\
0, & s^{\mathrm{pers}}_g < \eta_g \quad \text{(predicted failure)}.
\end{cases}
\end{equation}
The threshold can be selected using held-out successful examples to obtain a desired acceptance rate. Independent failures and near misses are then used to measure the false-positive rate. This keeps reward construction based on successful demonstrations only, while using separate labeled outcomes to validate its reliability.

The evaluation will compare positive-only IRR with a calibrated variant. 
In positive-only IRR, the reference bank and decision threshold are derived 
without using labeled failures; failures and near misses are reserved for 
independent evaluation. In the calibrated variant, a limited development set of labeled successes and failures is used either to select the decision 
threshold or to train a small reward head on the frozen representation. 
For example, threshold calibration can be performed by sweeping candidate 
values of $\eta_g$ on the calibration set and selecting the highest threshold that still achieves the desired success sensitivity while minimizing false positives.

\begin{figure}[htbp]
\centering
\begin{minipage}[t]{0.32\linewidth}
\includegraphics[width=\textwidth,height=5.5cm,keepaspectratio]{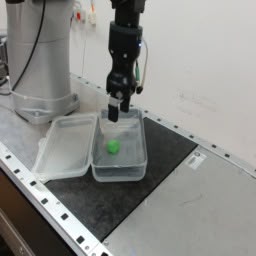}
\end{minipage}\hfill
\begin{minipage}[t]{0.32\linewidth}
\includegraphics[width=\textwidth,height=5.5cm,keepaspectratio]{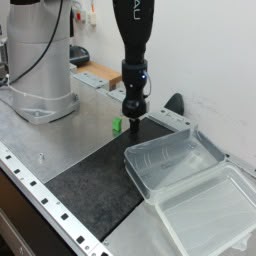}
\end{minipage}\hfill
\begin{minipage}[t]{0.32\linewidth}
\includegraphics[width=\textwidth,height=5.5cm,keepaspectratio]{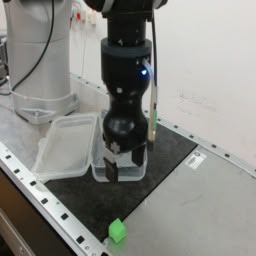}
\end{minipage}
\caption{\small Representative task outcomes. From left to right: successful completion, with the object released and resting inside the target container; grasp failure where the object is not securely picked; task failure where the object is dropped and not successfully recovered.}
\label{fig:outcomes}
\end{figure}

The additional labeling effort is recorded explicitly, allowing to assess when simple reference reuse is sufficient and when modest calibration improves reliability. Figure~\ref{fig:outcomes} shows representative outcome images for the distinction between visual resemblance and physical task completion.

\subsection{Policy improvement and control integration}
\label{sec:rl}
The first implementation will retain the imitation-trained VLA and train a bounded residual controller for Cartesian corrections. Residual RL provides a way for combining a prior controller with learned corrections~\citep{residual}. An off-policy actor--critic method such as Soft Actor-Critic (SAC) supplies a concrete stochastic learning mechanism without requiring likelihoods from the OFT regression head~\citep{sac}. In schematic form,
\begin{equation}
 a_t^{\mathrm{exec}}=\mathcal{S}\!\left(a_t^{0}+B\delta a_t\right),
 \quad a_t^{0}=\pi_0(o_t,p_t,g),\quad
 \delta a_t\sim\pi_\theta(\cdot\mid z_t,p_t,g,a_t^0).
 \label{eq:residual}
\end{equation}
where \(\delta a_t\) denotes the residual action correction produced by the learned policy \(\pi_\theta\). $B$ bounds and scales corrections in explicitly defined units and $\mathcal{S}$ is the fixed execution constraint layer. Equation~\eqref{eq:residual} applies to continuous Cartesian commands whereas initial gripper commands remain those of the base policy. Learning gripper timing would require an explicitly modeled discrete or hybrid action policy and is a subsequent extension. Residual improvement must consequently be reported as improvement of the combined controller, not as full VLA weight adaptation.

The learner stores commanded and executed actions, interventions, timestamps, and true termination causes. If the residual policy chooses $\delta a_t$, replay records that choice and the resulting transition under the fixed constraint layer, it must not relabel a clipped physical command as the sampled policy action. Reward features can reuse an existing policy encoder forward pass as long as images, preprocessing, and encoder weights are identical.

Full policy adaptation with a fixed reward-encoder snapshot is a later study. Likewise, dense progress shaping is an extension, because usually similarity to a final state does not increase monotonically along a valid trajectory. 

\begin{figure}[t]
\centering
\begin{tikzpicture}[
  box/.style={draw=black!60,rounded corners=2pt,align=center,font=\small,
    minimum height=1.1cm,text width=3.1cm,inner sep=5pt},
  arr/.style={-{Latex[length=2mm]},thick},
  lab/.style={font=\scriptsize,fill=white,inner sep=2pt}]
\node[box] (demo) at (0,0) {Successful\ demonstrations};
\node[box] (bank) at (5.2,0) {Task-specific\ reference bank};
\node[box] (obs) at (0,-2.1) {Camera\\ observations};
\node[box] (enc) at (5.2,-2.1) {Frozen VLA\ visual encoder};
\node[box] (reward) at (10.4,-2.1) {IRR score\ and validity checks};
\node[box] (robot) at (0,-4.4) {Industrial robot\ and workspace};
\node[box] (policy) at (5.2,-4.4) {VLA policy and\ bounded residual};
\node[box] (learn) at (10.4,-4.4) {Replay and\ policy improvement};
\draw[arr] (demo) -- node[lab,above]{encode} (bank);
\draw[arr] (obs) -- (enc);
\draw[arr] (enc) -- (reward);
\draw[arr] (bank.east) -| (reward.north);
\draw[arr] (enc) -- node[lab,right]{features} (policy);
\draw[arr] (reward) -- (learn);
\draw[arr] (learn) -- node[lab,above]{update} (policy);
\draw[arr] (policy) -- node[lab,above]{action} (robot);
\draw[arr] (robot) -- (obs);
\draw[arr,dashed] (demo.west) -- ++(-0.5,0) |- ([yshift=-9mm]policy.south)
  -- node[lab,right]{imitation initialization} (policy.south);
\end{tikzpicture}
\caption{\small Proposed learning loop. Reference endpoints use the same frozen encoder as policy observations. The policy also receives the task instruction and proprioception. Independent outcome audits assess success outside the IRR policy-update path.}
\label{fig:architecture}
\end{figure}
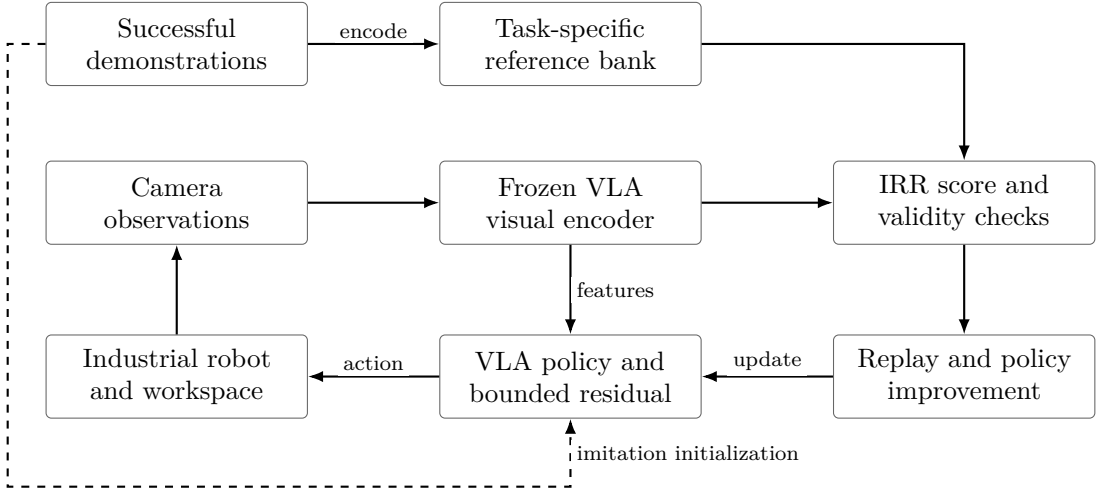

\section{Research questions and decisive experiments}
\label{sec:questions}

\textbf{RQ1 -- How effectively do existing policy representations support task evaluation?}
We hypothesize that a task-conditioned reference bank can distinguish successful endpoints from realistic near misses with a low false-positive rate. The evaluation will include correct placement, objects beside the container, objects still held above it, wrong-object placement, premature release, and partial occlusion. These comparisons will identify the features and reference structures that make existing representations useful for dependable outcome assessment.

\textbf{RQ2 -- How effectively does IRR support physical policy improvement?}
We hypothesize that residual RL using IRR feedback improves independently measured task success over the same imitation policy and control stack. Learning curves will show both IRR scores and independently verified outcomes. Their agreement is central to demonstrating useful improvement, divergence will reveal cases for refining the evaluator. A matched human-scored RL comparison will separate the contribution of reward provision from that of the learner and control configuration.

\textbf{RQ3 -- What efficiency benefits does reuse provide?}
We hypothesize that reusing the policy encoder and existing demonstrations reduces recurring human outcome-scoring time and the overhead of maintaining a separate evaluator. Evaluation will include reference construction, calibration, inference, audits, resets, and recovery. Measuring these costs together will establish the practical benefit per task and per verified successful cycle. Comparisons with dedicated reward models and further imitation training will place the benefit in the context of alternative uses of the same resources.

\textbf{RQ4 -- How can the approach support adaptation across tasks and conditions?}
Task-specific reference banks offer a simple mechanism for introducing new success conditions while retaining the encoder and evaluation interface. We will study variations in lighting, viewpoint, object appearance, spatial arrangement, and task identity. Adapting to a new task through a new demonstration bank will be distinguished from zero-shot transfer. Subsequent work can extend the approach to further robot embodiments and to phase-dependent or temporal references. Where a task requires additional views, proprioception, or tactile evidence, the resulting gains and costs will be evaluated explicitly.

\section{Evaluation design and evidence standards}
\label{sec:evaluation}
\subsection{Tasks, baselines, and ablations}
The first task reproduces the documented cube-to-container experiment. Two planned extensions introduce target-container placement with separation and placement into a fixture with visible pose constraints. These are laboratory tasks motivated by industrial handling. Natural-product sorting and precision assembly provide subsequent application directions, with additional acceptance measurements wherever appearance alone is inadequate.

The minimum physical comparison contains: (i) the fixed imitation policy (ii) the same residual RL learner with human binary outcome rewards (iii) that learner with strict positive-only IRR and (iv) the learner with one established external visual reward baseline selected during offline screening. Candidate external rewards include VIP/LIV-style goal distance or a pretrained robot reward VLM such as RoboReward. A calibrated IRR variant can also form an additional setting. All learning conditions start from the same policy checkpoint and receive matched robot-interaction experience. 

The main ablations should compare a single reference with a reference bank, global with spatial features, the two visual cores (DINOv2 and SigLIP) with their fusion, a single frame with persistence, and positive-only with labeled calibration. 

\subsection{Independent outcomes and metrics}
Before data collection, each task receives an operational success definition. For pick-and-place, this includes the correct object resting inside the intended container after release for a specified observation interval, within the deadline, with no disqualifying event. Human evaluators label synchronized video without seeing the used method or reward score. Such audit labels are used to evaluate IRR, not to reward its training episodes.

\begin{table}[ht]
\caption{\small Primary evaluation measures.}
\label{tab:metrics}
\small
\begin{tabularx}{\linewidth}{@{}L{3.0cm}Y@{}}
\toprule
Measure & Evaluation \\
\midrule
Task success rate &
Independently verified successes among all valid trials, report absolute differences with confidence intervals. \\

False-positive rate &
$\Pr(\hat y=1\mid y=0)$ on labeled failures, proposed target $\leq 2\%$. \\

Success sensitivity &
$\Pr(\hat y=1\mid y=1)$ on verified successes, proposed target $\geq 90\%$. \\

Learning efficiency &
Verified success versus robot interactions and learning time. \\

Human effort &
Time required for demonstrations, calibration, outcome scoring, audits, resets, and recovery. \\

Execution quality &
Completion time, retries, resets, safety interventions, and task-specific quality measures. \\

Reward cost &
Additional evaluator latency, memory, and computation, including extra views or frames. \\
\bottomrule
\end{tabularx}
\end{table}

At least 100 baseline trials will be used for diagnostic characterization. Confirmatory sample sizes depend on the expected effect and experimental replication.

\section{Scientific value and industrial relevance}
\label{sec:impact}

IRR advances a practical research position, that representations and demonstrations used to establish robot behavior can also support its evaluation and improvement. Reuse can extend the function of an existing VLA deployment while keeping the reward mechanism compact. The core design introduces task references and a scoring operation, with no separate learned evaluator or additional perception backbone. This makes the approach straightforward to relate to the robot's existing data and model configuration.

The potential benefit has several dimensions. Demonstrations can support both action learning and outcome evaluation, increasing the value of the collection effort. A shared representation can reduce the number of models that must be integrated, updated, and monitored. Task references provide a direct way to introduce new success conditions, and internally computed feedback can reduce repeated human scoring during learning. These are concrete opportunities arising from the architecture, their magnitude will be established through matched performance and effort measurements. The formulation also makes it possible to study how representation quality affects reward reliability and subsequent policy improvement.

Industrial handling offers a clear starting point. Variable-part manipulation, sorting into designated destinations, and placement into fixtures involve repeated adaptation to new objects and workspace conditions. For tasks with visually observable completion, demonstration-derived references offer a promising way to make reward provision easier to reuse across applications. The practical contribution will be assessed through commissioning effort, human time per verified successful cycle, throughput, and acceptance quality.

The proposed IRR learning loop will operate within independently safeguarded robot operation and task-specific quality assessment. Controller limits, stopping functions, and validated recovery procedures provide the operational framework for physical learning. Where acceptance depends on forces, hidden insertion depth, or product condition, the relevant measurements complement visual outcome evaluation. This maintains a clear role for each component while allowing the reward mechanism to remain focused on learning.

The direction also supports longer-term research into phase-aware progress representations, controlled expansion of reference banks, further robot embodiments, and direct VLA adaptation. Video world-model features, including V-JEPA~2~\citep{vjepa2}, provide a useful comparator for temporally structured tasks. The principle remains consistent across these extensions: make effective use of available representations and experience, and add capability where its contribution can be demonstrated.

\section{Conclusion}

Intrinsic Robot Rewarding puts existing VLA representations and successful demonstrations to work as a source of feedback for further learning. A task-specific reference bank and compact scoring mechanism provide a concrete route to internal outcome evaluation, with the potential to reduce recurring supervision and the integration of additional reward models.

The proposed work aims to connect reference-based reward generation with physical policy improvement. Its evaluation links reward reliability, verified task success, and total supervision effort, making the anticipated benefits measurable.

Our position is that self-evaluating and self-improving robots can benefit substantially from making fuller use of what they already have: pretrained representations, successful demonstrations, and experience from task execution. IRR develops this principle into a research direction for industrial manipulation. It offers a promising basis for extending imitation-trained behavior toward continued learning through the robot's own attempts.

\begingroup
\small
\interlinepenalty=10000
\setlength{\bibsep}{3pt}
\bibliographystyle{unsrtnat}
\bibliography{IRR_References}

@inproceedings{rt2,
  author = {Zitkovich, Brianna and Yu, Tianhe and Xu, Sichun and Xu, Peng and Xiao, Ted and Xia, Fei and Wu, Jialin and Wohlhart, Paul and Welker, Stefan and others},
  title = {{RT-2}: Vision-Language-Action Models Transfer Web Knowledge to Robotic Control},
  booktitle = {Proceedings of the 7th Conference on Robot Learning},
  series = {Proceedings of Machine Learning Research},
  volume = {229},
  pages = {2165--2183},
  year = {2023},
  url = {https://proceedings.mlr.press/v229/zitkovich23a.html}
}

@inproceedings{openvla,
  author = {Kim, Moo Jin and Pertsch, Karl and Karamcheti, Siddharth and Xiao, Ted and Balakrishna, Ashwin and Nair, Suraj and Rafailov, Rafael and Foster, Ethan P. and Sanketi, Pannag R. and Vuong, Quan and Kollar, Thomas and Burchfiel, Benjamin and Tedrake, Russ and Sadigh, Dorsa and Levine, Sergey and Liang, Percy and Finn, Chelsea},
  title = {{OpenVLA}: An Open-Source Vision-Language-Action Model},
  booktitle = {Proceedings of the 8th Conference on Robot Learning},
  series = {Proceedings of Machine Learning Research},
  volume = {270},
  pages = {2679--2713},
  year = {2025},
  url = {https://proceedings.mlr.press/v270/kim25c.html}
}

@inproceedings{pathak2017,
  author = {Pathak, Deepak and Agrawal, Pulkit and Efros, Alexei A. and Darrell, Trevor},
  title = {Curiosity-driven Exploration by Self-supervised Prediction},
  booktitle = {Proceedings of the 34th International Conference on Machine Learning},
  series = {Proceedings of Machine Learning Research},
  volume = {70},
  pages = {2778--2787},
  year = {2017},
  url = {https://proceedings.mlr.press/v70/pathak17a.html}
}

@misc{pipeline,
  author = {Elkhayat, Mohab and Almohamad, Mustafa and Schaffer, Tobias},
  title = {A Vision-Language-Action Pipeline for Robotic Pick-and-Place: System Design, {OpenVLA} Fine-Tuning, and Closed-Loop Evaluation},
  howpublished = {},
  year = {2026},
  note = {Manuscript in preparation, preprint to be made available},
}

@misc{dinov2,
  author = {Oquab, Maxime and Darcet, Timoth{\'e}e and Moutakanni, Th{\'e}o and Vo, Huy and Szafraniec, Marc and Khalidov, Vasil and Fernandez, Pierre and Haziza, Daniel and Massa, Francisco and El-Nouby, Alaaeldin and others},
  title = {{DINOv2}: Learning Robust Visual Features without Supervision},
  year = {2023},
  howpublished = {arXiv:2304.07193},
  url = {https://arxiv.org/abs/2304.07193}
}

@inproceedings{siglip,
  author = {Zhai, Xiaohua and Mustafa, Basil and Kolesnikov, Alexander and Beyer, Lucas},
  title = {Sigmoid Loss for Language Image Pre-Training},
  booktitle = {Proceedings of the IEEE/CVF International Conference on Computer Vision},
  year = {2023},
  url = {https://arxiv.org/abs/2303.15343}
}

@inproceedings{oft,
  author = {Kim, Moo Jin and Finn, Chelsea and Liang, Percy},
  title = {Fine-Tuning Vision-Language-Action Models: Optimizing Speed and Success},
  booktitle = {Robotics: Science and Systems},
  year = {2025},
  url = {https://arxiv.org/abs/2502.19645}
}

@misc{hilserl,
  author = {Luo, Jianlan and Xu, Charles and Wu, Jeffrey and Levine, Sergey},
  title = {Precise and Dexterous Robotic Manipulation via Human-in-the-Loop Reinforcement Learning},
  year = {2024},
  howpublished = {arXiv:2410.21845},
  url = {https://arxiv.org/abs/2410.21845}
}

@techreport{recap,
  author = {{Physical Intelligence}},
  title = {{$\pi_{0.6}^{*}$}: A {VLA} That Learns From Experience},
  institution = {Physical Intelligence},
  year = {2025},
  type = {Technical report},
  url = {https://www.pi.website/download/pistar06.pdf}
}

@inproceedings{vip,
  author = {Ma, Yecheng Jason and Sodhani, Shagun and Jayaraman, Dinesh and Bastani, Osbert and Kumar, Vikash and Zhang, Amy},
  title = {{VIP}: Towards Universal Visual Reward and Representation via Value-Implicit Pre-Training},
  booktitle = {International Conference on Learning Representations},
  year = {2023},
  url = {https://arxiv.org/abs/2210.00030}
}

@inproceedings{liv,
  author = {Ma, Yecheng Jason and Kumar, Vikash and Zhang, Amy and Bastani, Osbert and Jayaraman, Dinesh},
  title = {{LIV}: Language-Image Representations and Rewards for Robotic Control},
  booktitle = {Proceedings of the 40th International Conference on Machine Learning},
  series = {Proceedings of Machine Learning Research},
  volume = {202},
  pages = {23301--23320},
  year = {2023},
  url = {https://proceedings.mlr.press/v202/ma23b.html}
}

@inproceedings{roboclip,
  author = {Sontakke, Sumedh and Zhang, Jesse and Arnold, S{\'e}b and Pertsch, Karl and B{\i}y{\i}k, Erdem and Sadigh, Dorsa and Finn, Chelsea and Itti, Laurent},
  title = {{RoboCLIP}: One Demonstration is Enough to Learn Robot Policies},
  booktitle = {Advances in Neural Information Processing Systems},
  volume = {36},
  year = {2023},
  url = {https://arxiv.org/abs/2310.07899}
}

@misc{vlmrewards,
  author = {Baumli, Kate and Baveja, Satinder and Behbahani, Feryal and Chan, Harris and Comanici, Gheorghe and Flennerhag, Sebastian and Gazeau, Maxime and Holsheimer, Kristian and Horgan, Dan and Laskin, Michael and others},
  title = {Vision-Language Models as a Source of Rewards},
  year = {2023},
  howpublished = {arXiv:2312.09187},
  note = {Cited version: v3, revised July 2024},
  url = {https://arxiv.org/abs/2312.09187v3}
}

@inproceedings{rlvlmf,
  author = {Wang, Yufei and Sun, Zhanyi and Zhang, Jesse and Xian, Zhou and Biyik, Erdem and Held, David and Erickson, Zackory},
  title = {{RL-VLM-F}: Reinforcement Learning from Vision Language Foundation Model Feedback},
  booktitle = {Proceedings of the 41st International Conference on Machine Learning},
  series = {Proceedings of Machine Learning Research},
  volume = {235},
  pages = {51484--51501},
  year = {2024},
  url = {https://proceedings.mlr.press/v235/wang24bn.html}
}

@inproceedings{goalladder,
  author = {Zakharov, Alexey and Whiteson, Shimon},
  title = {{GoalLadder}: Incremental Goal Discovery with Vision-Language Models},
  booktitle = {Advances in Neural Information Processing Systems},
  year = {2025},
  url = {https://arxiv.org/abs/2506.16396}
}

@misc{srpo,
  author = {Fei, Senyu and Wang, Siyin and Ji, Li and Li, Ao and Zhang, Shiduo and Liu, Liming and Hou, Jinlong and Gong, Jingjing and Zhao, Xianzhong and Qiu, Xipeng},
  title = {{SRPO}: Self-Referential Policy Optimization for Vision-Language-Action Models},
  year = {2025},
  howpublished = {arXiv:2511.15605},
  note = {Cited version: v2, 30 November 2025},
  url = {https://arxiv.org/abs/2511.15605v2}
}

@misc{vjepa2,
  author = {Assran, Mido and Bardes, Adrien and Fan, David and others},
  title = {{V-JEPA 2}: Self-Supervised Video Models Enable Understanding, Prediction and Planning},
  year = {2025},
  howpublished = {arXiv:2506.09985},
  url = {https://arxiv.org/abs/2506.09985}
}

@misc{roboreward,
  author = {Lee, Tony and Wagenmaker, Andrew and Pertsch, Karl and Liang, Percy and Levine, Sergey and Finn, Chelsea},
  title = {{RoboReward}: General-Purpose Vision-Language Reward Models for Robotics},
  year = {2026},
  howpublished = {arXiv:2601.00675},
  note = {Cited version: v2, 8 January 2026},
  url = {https://arxiv.org/abs/2601.00675v2}
}

@misc{largereward,
  author = {Wu, Yanru and Yuan, Weiduo and Qi, Ang and Guizilini, Vitor and Mao, Jiageng and Wang, Yue},
  title = {Large Reward Models: Generalizable Online Robot Reward Generation with Vision-Language Models},
  year = {2026},
  howpublished = {arXiv:2603.16065},
  note = {Cited version: v2, 22 March 2026},
  url = {https://arxiv.org/abs/2603.16065v2}
}

@inproceedings{residual,
  author = {Johannink, Tobias and Bahl, Shikhar and Nair, Ashvin and Luo, Jianlan and Kumar, Avinash and Loskyll, Matthias and Aparicio Ojea, Juan and Solowjow, Eugen and Levine, Sergey},
  title = {Residual Reinforcement Learning for Robot Control},
  booktitle = {IEEE International Conference on Robotics and Automation},
  year = {2019},
  url = {https://arxiv.org/abs/1812.03201}
}

@inproceedings{sac,
  author = {Haarnoja, Tuomas and Zhou, Aurick and Abbeel, Pieter and Levine, Sergey},
  title = {Soft Actor-Critic: Off-Policy Maximum Entropy Deep Reinforcement Learning with a Stochastic Actor},
  booktitle = {Proceedings of the 35th International Conference on Machine Learning},
  series = {Proceedings of Machine Learning Research},
  volume = {80},
  pages = {1861--1870},
  year = {2018},
  url = {https://proceedings.mlr.press/v80/haarnoja18b.html}
}
\endgroup
\end{document}